\documentclass[runningheads]{llncs}

\usepackage{eccv}

\usepackage{eccvabbrv}
\usepackage{multirow}

\usepackage{graphicx}
\usepackage{booktabs}

\usepackage[accsupp]{axessibility}  

\usepackage{hyperref}
\usepackage{url}
\usepackage{orcidlink}

\begin{document}

\title{Spatial-Frequency Gated Swin Transformer for Cross-Sensor Remote Sensing Super-Resolution}

\titlerunning{SFG-SwinSR for Cross-Sensor Remote Sensing Super-Resolution}

\author{Md Aminur Hossain\inst{1}\orcidlink{0009-0003-6357-7480} \and
Ayush V. Patel\inst{2}\orcidlink{0000-0002-8921-7091} \and
Sanjay K. Singh\inst{1} \and
Yogesh Jethani\inst{3} \and
Biplab Banerjee\inst{2}\orcidlink{0000-0001-8371-8138}}

\authorrunning{M.A. Hossain et al.}

\institute{Space Applications Centre, ISRO, Ahmedabad, India \and
Centre of Studies in Resources Engineering, Indian Institute of Technology Bombay, India \and
GLS University, Ahmedabad, India\\
\email{\{md.aminurhossain,getbiplab,ayu020503\}@gmail.com}}

\maketitle

\begin{abstract}
Remote sensing single-image super-resolution aims to generate high-resolution imagery from low-resolution observations while preserving fine structures such as roads, building boundaries, field edges, and land-cover transitions. Swin Transformer-based models, including Swin2SR, provide strong spatial context modeling through shifted-window self-attention, but their feed-forward networks remain generic channel-mixing modules that do not explicitly distinguish low-frequency structure from residual details. We propose SFG-SwinSR, which replaces the standard Swin2SR feed-forward network with a lightweight Spatial-Frequency Gated Feed-Forward Network. The module estimates a smoothed feature component through a depthwise low-pass branch, derives residual details by subtraction, refines them spatially, and adaptively reinjects useful details through a bottleneck gate. Experiments on the real cross-sensor SEN2VEN$\mu$S, OLI2MSI, and SEN2NAIP benchmarks, together with an auxiliary synthetic SpaceNet Challenge 3 setting, show consistent improvements across most evaluation settings and competitive performance against recent Swin-based baselines. The results indicate that spatial-frequency transformation within transformer feed-forward networks provides an effective lightweight inductive bias for structure-aware cross-sensor remote sensing super-resolution. Source code is available at \url{https://github.com/aminurhossain/SFG-SwinSR}.

\end{abstract}
\section{Introduction}
\label{sec:intro}

High-resolution remote sensing imagery is essential for accurate Earth observation, supporting applications such as urban mapping, precision agriculture, environmental monitoring, infrastructure analysis, and disaster management. Nevertheless, the spatial resolution of satellite images is limited by sensor characteristics, imaging optics, orbital geometry, atmospheric conditions, and sensor-induced artifacts such as stripe noise~\cite{hossain2025weighted}. These limitations, along with blur, downsampling, geometric misregistration, and radiometric differences across sensors, suppress fine spatial structures such as narrow roads, building outlines, field boundaries, and land-cover transitions. Therefore, remote sensing single-image super-resolution (RS-SISR) remains an important and active research problem, particularly in cross-sensor settings where low-resolution and high-resolution observations are acquired by different imaging systems. Early deep learning approaches for RS-SISR predominantly relied on CNNs (Convolutional Neural Networks)~\cite{lei2017localglobal,ren2021dual}, but there is a current trend toward the use of transformer-based architectures such as Swin Transformer~\cite{tu2022swcgan,kang2024efficient,rossi2025swin2mose} because of their ability to capture long-range dependencies efficiently. In addition, recent studies on real-world and cross-sensor RS-SISR emphasize the need for degradation-aware training, careful benchmark design, and realistic evaluation protocols~\cite{zhang2025swinfr}.

In remote sensing imagery, high-frequency cues are important for object delineation and the preservation of land-cover boundaries. Accurate reconstruction of edges, textures, roads, rooftops, and field boundaries is therefore useful for visual interpretation and downstream analysis~\cite{qi2026survey,dong2022kanet}. At the same time, single-image super-resolution is an ill-posed problem: the true high-resolution signal cannot be uniquely recovered from a single low-resolution observation without additional assumptions. Therefore, the practical goal of learning-based RS-SR is to generate structurally consistent and sensor-plausible high-resolution imagery conditioned on the observed low-resolution input. Although Transformer-based models improve contextual modeling, their feature transformation stages are typically implemented using generic feed-forward networks (FFNs). These FFNs do not explicitly distinguish between low-frequency structural content and residual detail features, which can limit structure-aware reconstruction in remote sensing super-resolution.

To address this limitation, we propose a Spatial-Frequency Gated Swin Transformer for cross-sensor remote sensing super-resolution. We modify the Swin2SR~\cite{conde2022swin2sr} transformer block by replacing the standard FFN with a lightweight Spatial-Frequency Gated Feed-Forward Network (SFG-FFN). The proposed module estimates low-frequency content using a depthwise blur operator, derives residual detail features by subtraction, refines the residual through a lightweight spatial branch, and adaptively injects useful details through a bottleneck gating mechanism. This localized architectural modification preserves the shifted-window
attention mechanism and long-range dependency modeling capability of
Swin2SR while introducing a feature-space spatial-frequency inductive bias
inside the FFN. Experiments on real cross-sensor benchmarks, including SEN2VEN$\mu$S, OLI2MSI, and SEN2NAIP, together with a synthetic SpaceNet Challenge 3 setting, show that the proposed method provides consistent improvements across most evaluation settings and remains competitive with recent Swin-based remote sensing SR baselines. The synthetic SpaceNet Challenge 3 setting is included because many high-resolution satellite products lack corresponding native LR-HR image pairs; therefore, controlled degradation provides a practical auxiliary protocol for evaluating SR behavior under known downsampling conditions.

The main contributions of this work are summarized as follows:
\begin{enumerate}
\item We propose a Spatial-Frequency Gated Feed-Forward Network (SFG-FFN) that introduces low-frequency estimation and residual-detail modulation inside Swin Transformer blocks for cross-sensor remote sensing super-resolution.
\item We introduce a lightweight residual-detail refinement and an adaptive bottleneck-gating mechanism that modifies only the FFN component while preserving the original shifted-window attention design of Swin2SR.
\item We evaluate the proposed method on multiple real cross-sensor remote sensing SR benchmarks, including SEN2VEN$\mu$S, OLI2MSI, and SEN2NAIP, along with a synthetic SpaceNet Challenge 3 setting, and compare it with CNN-based, Transformer-based, and recent Swin-based SR baselines.
\end{enumerate}

\section{Related Work}

\textbf{CNN-based Image Super-Resolution:} Single-image super-resolution (SR) has been widely explored using convolutional neural networks (CNNs). The seminal study of SRCNN~\cite{dong2014learning} showed that end-to-end-trained mappings significantly surpass traditional interpolation methods. Later, models such as VDSR~\cite{kim2016accurate} and EDSR~\cite{lim2017enhanced} leveraged deep network architectures and residual learning paradigms to enhance reconstruction performance. Further progress was made using RDN~\cite{zhang2018residual} and RCAN~\cite{zhang2018rcan}, which harnessed dense hierarchical reuse of features and channel attention, respectively. In particular, RCAN illustrated that the skip connection of rich low-frequency details via residual connections enables the network to dedicate more resources for modeling informative high-frequency details. This observation is closely related to the motivation of our proposed spatial-frequency feature transformation. Although CNN-based models are highly effective, their feature extraction is primarily governed by local convolutional operations, which can limit their ability to model long-range spatial dependencies in large and structurally complex satellite scenes.

\textbf{Remote-Sensing Image Super-Resolution:} The field of remote sensing SR has seen advancements through various deep learning models, from task-oriented pipelines to more advanced architectures. Various surveys have been carried out in this respect, which show that several challenges remain, namely issues associated with preserving textures, consistency of geometric features, and domain discrepancy between the synthetic degradation models used and the actual sensing process~\cite{fernandez2017singleframe,qi2026survey}. These issues are particularly important in cross-sensor SR, where LR and HR observations are acquired from different sensors and may exhibit geometric misregistration, radiometric mismatch, and different acquisition conditions~\cite{michel2025revisiting}. Some notable works in this regard include the local-global fusion approach by Lei et al.~\cite{lei2017localglobal}, the modified dual-luminance architecture introduced by Ren et al.~\cite{ren2021dual}, the DSen2 method of Lanaras et al.~\cite{lanaras2018dsen2} specifically designed for Sentinel-2 imagery, and SWCGAN~\cite{tu2022swcgan}, an adversarial architecture based on Swin Transformer. For cross-sensor remote sensing SR, Qiu et al.~\cite{qiu2023crosssensor} introduced an edge-guided attention-based network to improve the learning of structural and high-frequency features across different sensors.

\textbf{Transformer-based Super-Resolution:} Transformer networks have become popular recently for image restoration tasks because of their ability to learn both local and non-local interactions using shifted-window self-attention mechanisms introduced in the Swin Transformer architecture. As a result, SwinIR~\cite{liang2021swinir} and Swin2SR~\cite{conde2022swin2sr} architectures have set strong benchmarks in image restoration and super-resolution. Concerning transformer-based SR in the context of remote sensing, recent years have seen research along various lines of development, including hybrid convolutional and transformer-based generator models~\cite{tu2022swcgan}, real-world degradation-aware architectures such as KANet~\cite{dong2022kanet}, efficiency-oriented Swin models like ESTNet~\cite{kang2024efficient}, and mixture-of-expert feed-forward layers such as Swin2-MoSE~\cite{rossi2025swin2mose}. Among these approaches, Swin2SR is of particular interest to our work because it provides a strong and reproducible Swin-based restoration backbone while retaining a standard transformer feed-forward network. Nevertheless, the FFN in each transformer block remains a generic two-layer MLP and does not explicitly model the separation between low-frequency structural content and residual detail features.

\textbf{Frequency-Aware Restoration:} The reconstruction of high-frequency cues such as edges, textures, and fine structures is essential in remote sensing interpretation. It is widely reported that preserving such information remains challenging under sensor blur, aliasing, misregistration, and cross-sensor radiometric differences~\cite{qi2026survey,michel2025revisiting}. To address these challenges, existing methods rely on perceptual losses, multi-branch refinement blocks, degradation-aware training processes~\cite{dong2022kanet}, edge-guided attention~\cite{qiu2023crosssensor}, or frequency-aware designs. For example, FDFNet~\cite{zhao2025fdfnet} decouples low- and high-frequency components for remote sensing image SR and introduces a frequency-aware transformer module, while EFATSR~\cite{song2025efatsr} aggregates frequency-domain features for natural image SR using a transformer-based design. These works show that frequency-aware modeling is useful for SR, but they mainly introduce frequency processing through dedicated frequency branches, attention modules, or network-level fusion designs.

In contrast, our objective is more localized and lightweight: we focus on the feed-forward transformation inside Swin Transformer blocks. Instead of replacing the attention mechanism or adding a separate frequency-fusion network, the proposed SFG-FFN modifies the FFN branch by estimating low-frequency features through depthwise blurring, deriving residual detail features by subtraction, refining them with a lightweight spatial branch, and adaptively reinjecting useful details through a bottleneck gate. This design preserves the original shifted-window attention structure of Swin2SR while introducing a spatial-frequency inductive bias directly into the transformer FFN.

\section{Proposed Method}

\subsection{Overall Architecture}

Given a low-resolution remote sensing image $\mathbf{I}_{LR}$, single-image
super-resolution aims to learn a mapping (Eq.~\ref{eq:problem}) that generates
a high-resolution estimate while preserving large-scale structure, radiometric
consistency, and fine spatial details. Our method builds on Swin2SR by modifying
the feed-forward network (FFN) within each Swin Transformer block, encouraging
a feature-space decomposition into a smoothed component and a residual-detail
component.

\begin{equation}
\hat{\mathbf{I}}_{HR}
=
\mathcal{F}_{\theta}(\mathbf{I}_{LR}),
\label{eq:problem}
\end{equation}

\begin{figure*}[t]
\centering
\includegraphics[width=\linewidth]{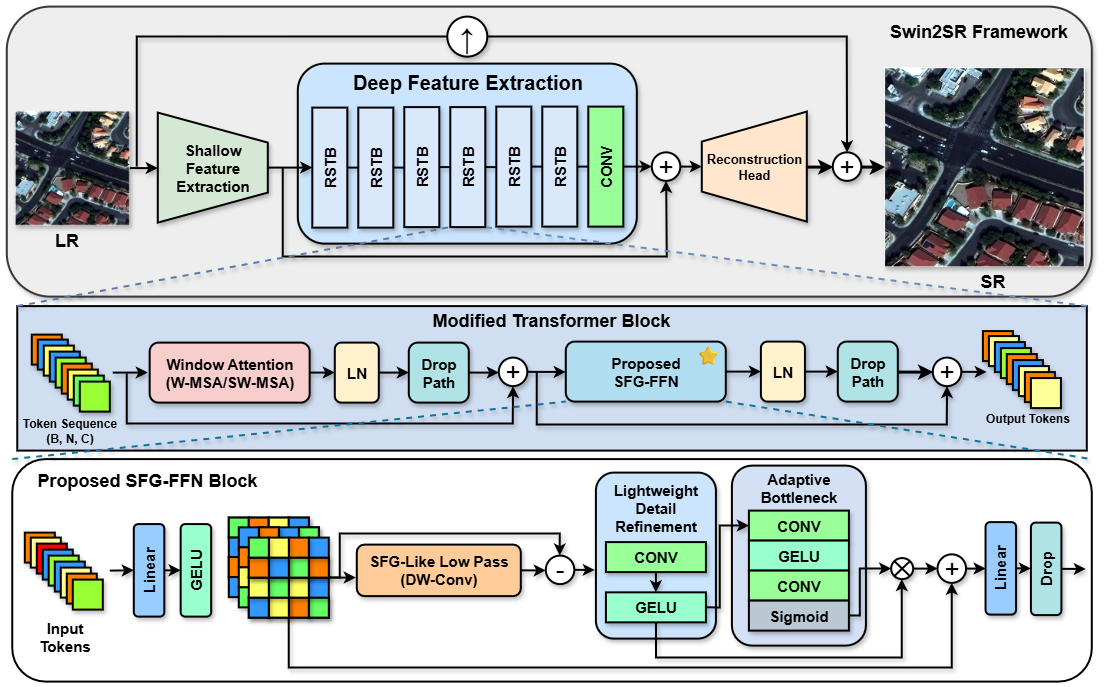}
\caption{Architecture of the proposed SFG-SwinSR. Each transformer block
replaces the standard FFN with an SFG-FFN that performs feature-space
low-pass estimation, residual-detail refinement, and adaptive gated fusion
for structure-aware remote sensing super-resolution.}
\label{fig:main_arch}
\end{figure*}

The proposed \textbf{SFG-SwinSR} builds upon the Swin2SR framework for
cross-sensor remote sensing super-resolution. Given a low-resolution satellite
image
$\mathbf{I}_{LR}\in\mathbb{R}^{B\times b\times H\times W}$,
the model generates a high-resolution estimate
$\hat{\mathbf{I}}_{HR}\in
\mathbb{R}^{B\times b\times sH\times sW}$,
where $B$ is the batch size, $b$ is the number of spectral bands, and $s$
is the upscaling factor. In our experiments, $s$ varies across datasets
according to the corresponding cross-sensor setting.

The full architecture pipeline shown in Figure~\ref{fig:main_arch} is composed of three major steps: a shallow convolutional embedding layer that projects the input data into a $C$-dimensional feature space, followed by a deep feature extraction step that includes six Swin Transformer stages with each containing six transformer layers, and finally a pixel upsampling reconstruction head~\cite{shi2016pixelshuffle}. The critical modification made in this paper is replacing the traditional two-layer multilayer perceptron feed-forward network in each Swin2SR transformer layer with the novel \textbf{Spatial-Frequency Gated Feed-Forward Network (SFG-FFN)}.

\subsection{Spatial-Frequency Gated Feed-Forward Network (SFG-FFN)}

Given the input token sequence
$\mathbf{x} \in \mathbb{R}^{B \times HW \times C}$,
the module first applies a linear projection to expand the channel dimension using GELU (Gaussian Error Linear Unit) activation:

\begin{equation}
\mathbf{z}
=
\text{GELU}\left(
\mathbf{x}\mathbf{W}_1^{\top}+\mathbf{b}_1
\right)
\in \mathbb{R}^{B \times HW \times C_h}
\label{eq:expansion}
\end{equation}

where $\mathbf{W}_1 \in \mathbb{R}^{C_h \times C}$,
$\mathbf{b}_1 \in \mathbb{R}^{C_h}$, and
$C_h = \lfloor C \cdot r \rfloor$ is the hidden dimension determined by
the MLP expansion ratio $r = 2.0$. With $C = 180$, this gives $C_h = 360$.

The expanded token sequence is reshaped into a 2D feature map:

\begin{equation}
\mathbf{F} = \text{Reshape}(\mathbf{z})
\in \mathbb{R}^{B \times C_h \times H \times W}
\label{eq:reshape}
\end{equation}

The low-frequency component estimation process employs a depthwise convolution
whose initial weights are equivalent to those of a uniform low-pass filter:

\begin{equation}
\mathbf{F}_{LF} = \text{DWConv}_{k \times k}(\mathbf{F}),
\quad w_{i,j} = \frac{1}{k^2}, \quad k = 5
\label{eq:lowfreq}
\end{equation}
The kernel weights are initialized as $w_{i,j} = 1/25$ for all $i,j \in \{1,\ldots,5\}$, providing a simple low-pass prior~\cite{gonzalez2002digital} while remaining trainable.

The residual detail feature is calculated through the difference between the input feature map and the estimated low-frequency component as follows:

\begin{equation}
\mathbf{F}_{HF} = \mathbf{F} - \mathbf{F}_{LF}
\in \mathbb{R}^{B \times C_h \times H \times W}
\label{eq:highfreq}
\end{equation}

Here, $\mathbf{F}_{LF}$ and $\mathbf{F}_{HF}$ denote low- and high-spatial-variation components in the learned feature space, rather than guaranteed physical frequency components of the original HR scene. This distinction is important because single-image SR is inherently ill-posed, especially in cross-sensor settings.

The residual detail feature is then further improved by a lightweight spatial refinement branch as follows:

\begin{equation}
\tilde{\mathbf{F}}_{HF} =
\text{GELU}(\text{DWConv}_{3 \times 3}(\mathbf{F}_{HF}))
\in \mathbb{R}^{B \times C_h \times H \times W}
\label{eq:refinement}
\end{equation}

The adaptive bottleneck gate~\cite{hu2018squeeze} determines the level of influence of residual detail features:

\begin{equation}
\mathbf{G}
=
\sigma\left(
\text{Conv}_{1\times1}^{C_g\rightarrow C_h}
\left(
\text{GELU}\left(
\text{Conv}_{1\times1}^{C_h\rightarrow C_g}
\left(\tilde{\mathbf{F}}_{HF}\right)
\right)
\right)
\right)
\in \mathbb{R}^{B \times C_h \times H \times W}
\label{eq:gate}
\end{equation}

where $C_g=\max(\lfloor C_h/\rho\rfloor,16)$ with reduction factor
$\rho=8$~\cite{hu2018squeeze}, giving $C_g=45$.

The refined residual details are added back to the original feature map through a lightweight element-wise gate~\cite{he2016deep}, where $\odot$ denotes element-wise multiplication:

\begin{equation}
\mathbf{F}_{out} = \mathbf{F} +
\mathbf{G} \odot \tilde{\mathbf{F}}_{HF}
\label{eq:fusion}
\end{equation}

The output feature map is reshaped into tokens, projected back to the
original channel dimension, and optionally passed through dropout:

\begin{equation}
\mathbf{y}
=
\operatorname{Dropout}\left(
\operatorname{Reshape}(\mathbf{F}_{out})\mathbf{W}_2^{\top}
+\mathbf{b}_2
\right)
\in\mathbb{R}^{B\times HW\times C},
\label{eq:projection}
\end{equation}

where $\mathbf{W}_2\in\mathbb{R}^{C\times C_h}$ and
$\mathbf{b}_2\in\mathbb{R}^{C}$. The token sequence $\mathbf{y}$
is the output of the SFG-FFN.

\subsection{Swin Transformer Block with Spatial-Frequency Gated FFN}

Swin2SR alternates window-based multi-head self-attention (W-MSA) and
shifted-window multi-head self-attention (SW-MSA), followed by a
feed-forward transformation. For input tokens
$\mathbf{x}\in\mathbb{R}^{B\times HW\times C}$, the modified block is
expressed as:

\begin{equation}
\mathbf{x}'
=
\mathbf{x}
+
\text{DropPath}\left(
\text{LN}_1\left(
\text{W/SW-MSA}(\mathbf{x})
\right)
\right)
\label{eq:attn_residual}
\end{equation}

\begin{equation}
\mathbf{x}_{out} = \mathbf{x}' +
\text{DropPath}(\text{LN}_2(\text{SFG-FFN}(\mathbf{x}')))
\label{eq:ffn_residual}
\end{equation}

Here, $\text{LN}_1$ and $\text{LN}_2$ denote layer normalization~\cite{ba2016layer}, and DropPath denotes stochastic depth regularization~\cite{huang2016deep}. The SFG-FFN replaces the standard feed-forward network in each block with a spatial-frequency gated module, introducing a trainable feature-space low-pass/residual decomposition within the transformer FFN.

\subsection{Training Objective}

The model is trained using a detail-guided composite loss, summarized in
Table~\ref{tab:loss_functions}, to jointly encourage pixel-level accuracy,
structural fidelity, edge preservation, and frequency-domain consistency. The
overall objective consists of four complementary components: pixel-wise L1 loss,
SSIM loss, edge guidance loss, and frequency guidance loss.

In our experiments, the weights are set to $\lambda_{L1}=1.0$, $\lambda_{SSIM}=0.1$, $\lambda_{edge}=0.1$, and $\lambda_{freq}=0.05$. These weights are kept fixed across datasets to avoid dataset-specific tuning. The edge and frequency losses provide additional structural and
frequency-domain guidance, while their incremental contributions are
examined in the ablation study.

\begin{table*}[ht]
\centering
\caption{Detail-guided loss components. $\hat{I}$ and $I$ denote the SR and HR images, $\nabla$ is the spatial gradient operator, $\mathcal{F}(\cdot)$ is the 2D Fourier transform, $|\mathcal{F}(\cdot)|$ denotes the amplitude spectrum, and $\lambda_{\text{L1}}$, $\lambda_{\text{SSIM}}$, $\lambda_{\text{Edge}}$, and $\lambda_{\text{Freq}}$ are weighting hyperparameters.}
\label{tab:loss_functions}
\small
\setlength{\tabcolsep}{15pt}
\renewcommand{\arraystretch}{1.5}
\begin{tabular}{@{} l l @{}}
\toprule
\textbf{Component} & \textbf{Mathematical Formulation} \tabularnewline
\midrule
Pixel-wise (L1) & $\displaystyle \mathcal{L}_{\text{L1}} = \frac{1}{N} \sum_{i=1}^{N} \| \hat{I}_i - I_i \|_1$ \tabularnewline
\addlinespace
Structural (SSIM) & $\displaystyle \mathcal{L}_{\text{SSIM}} = 1 - \text{SSIM}(\hat{I}, I)$ \tabularnewline
\addlinespace
Edge Guidance & $\displaystyle \mathcal{L}_{\text{Edge}} = \frac{1}{N} \sum_{i=1}^{N} \| \nabla \hat{I}_i - \nabla I_i \|_1$ \tabularnewline
\addlinespace
Frequency Guidance & $\displaystyle \mathcal{L}_{\text{Freq}} = \frac{1}{N} \sum_{i=1}^{N} \left\| |\mathcal{F}(\hat{I}_i)| - |\mathcal{F}(I_i)| \right\|_1$ \tabularnewline
\addlinespace
\textbf{Total Objective} & $\displaystyle \mathcal{L}_{\text{Total}} = \lambda_{L1} \mathcal{L}_{\text{L1}} + \lambda_{SSIM} \mathcal{L}_{\text{SSIM}} + \lambda_{edge} \mathcal{L}_{\text{Edge}} + \lambda_{freq} \mathcal{L}_{\text{Freq}}$ \tabularnewline
\bottomrule
\end{tabular}
\end{table*}

\section{Experimental Setup}

\subsection{Dataset Preparation}

We evaluate the proposed method on three real cross-sensor remote sensing super-resolution benchmarks, namely SEN2VEN$\mu$S, OLI2MSI, and SEN2NAIP, together with a synthetic SpaceNet Challenge 3 setting. The real cross-sensor datasets are used as the primary evaluation setting, while SpaceNet is included as a controlled synthetic benchmark.

\textbf{SEN2VEN$\mu$S Dataset}. We use SEN2VEN$\mu$S version 2~\cite{michel2022sen2venus}, which pairs Sentinel-2 observations at 10~m and 20~m resolution with VEN$\mu$S reference images at 5~m resolution acquired within a 30-minute time window. Following our computational budget, we use a fixed randomly selected subset of 20,000 samples from the full dataset. This subset is split into 18,000 training samples, of which 10\% are used for validation, and 2,000 samples for final testing. We evaluate both $\times2$ and $\times4$ super-resolution settings.

\textbf{OLI2MSI Dataset}. OLI2MSI~\cite{wang2021multisensor} is a multi-sensor benchmark containing paired Landsat-8 OLI and Sentinel-2 MSI images. Landsat-8 OLI images at 30~m resolution are used as LR inputs, and Sentinel-2 MSI images at 10~m resolution are used as HR references, resulting in a $\times3$ super-resolution task. The dataset contains 5,325 aligned multispectral image pairs. We follow the official split with 5,225 training pairs and 100 test pairs, and reserve 10\% of the training set for validation.

\textbf{SEN2NAIP Dataset}. SEN2NAIP~\cite{aybar2024sen2naip} pairs Sentinel-2 imagery with high-resolution aerial imagery from the National Agriculture Imagery Program (NAIP). Sentinel-2 RGB-NIR images at 10~m resolution serve as LR inputs, while NAIP images at 2.5~m resolution are used as HR references, corresponding to a $\times4$ setting. We use the standard split of 2,280 training pairs and 571 test pairs, with 10\% of the training set used for validation.

\textbf{SpaceNet Challenge 3 (WorldView-3)}. We also include a synthetic SR setting using RGB image products from SpaceNet Challenge 3~\cite{vanetten2018spacenet}. The dataset contains high-resolution WorldView-3 imagery from four cities: Las Vegas, Paris, Shanghai, and Khartoum. Although the original benchmark provides road-network annotations, we use only the satellite image chips for super-resolution evaluation. Since native LR-HR image pairs are not provided, LR images are generated from HR images using a controlled degradation process consisting of blur, downsampling, and noise. We obtain 12,144 valid LR-HR image pairs, including 10,261 training samples and 1,883 test samples. This dataset is used as an auxiliary synthetic benchmark.

During training, aligned LR-HR crops are extracted according to the scale factor of each dataset. For SpaceNet Challenge 3, we use $64 \times 64$ LR and $128 \times 128$ HR patches for $\times2$ SR. For SEN2VEN$\mu$S, $\times2$ SR uses $128 \times 128$ LR and $256 \times 256$ HR patches, while $\times4$ SR uses $64 \times 64$ LR and $256 \times 256$ HR patches. For OLI2MSI and SEN2NAIP, aligned crops are extracted following their native scale factors of $\times3$ and $\times4$, respectively. No test-time augmentation is used, and the same preprocessing protocol is applied to all compared methods.

\begin{figure}[t]
    \centering
    \includegraphics[width=\linewidth]{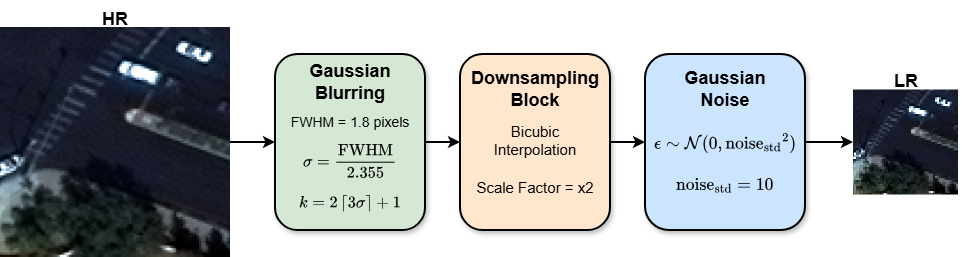}
    \caption{Controlled HR-to-LR degradation pipeline used to generate
    synthetic SpaceNet image pairs.}
    \label{fig:hr_to_lr}
\end{figure}

\textbf{LR Image Simulation for SpaceNet.}
Figure~\ref{fig:hr_to_lr} shows the controlled degradation pipeline,
adapted from~\cite{dong2022kanet}, used to generate low-resolution images
from 30-cm WorldView-3 imagery. Given an HR image $\mathbf{I}_{HR}$,
the LR image is generated as
\begin{equation}
\mathbf{I}_{LR} =
\mathcal{N}\!\left(
\mathcal{D}\!\left(
\mathcal{B}(\mathbf{I}_{HR})
\right)
\right),
\label{eq:degradation}
\end{equation}
where $\mathcal{B}(\cdot)$ applies Gaussian blurring with an FWHM of
$1.8$ pixels ($\sigma\approx0.764$, $7\times7$ kernel),
$\mathcal{D}(\cdot)$ denotes bicubic downsampling by a factor of two, and
$\mathcal{N}(\cdot)$ adds zero-mean Gaussian noise with a standard deviation
of 10 DN. The output is clipped to the valid range $[0,2047]$.

\subsection{Implementation Details}

All models are implemented in PyTorch and trained end-to-end on a single NVIDIA A100-PCIE-40GB GPU. Unless otherwise stated, we optimize the networks using AdamW with an initial learning rate of $2\times10^{-4}$ and weight decay of $1\times10^{-4}$. Training is conducted for 200 epochs with a batch size of 16 using a cosine annealing schedule with a minimum learning rate of $1\times10^{-6}$. For reproducibility, a fixed random seed of 42 is used for data shuffling and weight initialization.

SFG-SwinSR adopts the Swin2SR backbone and modifies only the feed-forward network (FFN) within each transformer block. The model is configured with window size 8, embedding dimension $C=180$, six stages with depths $[6,6,6,6,6,6]$ for a total of 36 transformer blocks, six attention heads per block, and an MLP expansion ratio of $r=2.0$. In each block, the standard FFN is replaced by the proposed SFG-FFN with depthwise blur kernel size $k=5$ and reduction ratio $\rho=8$. All other components, including shifted-window partitioning and the reconstruction head, are kept identical to Swin2SR to isolate the effect of the proposed frequency-gated FFN design.

We compare SFG-SwinSR with bicubic interpolation, CNN-based baselines
such as EDSR and RCAN, and Swin-based transformer baselines including
Swin2SR and Swin2-MoSE.

PSNR, SSIM, and MAE are computed over all reconstructed bands of each
test image after mapping the predictions and reference images to the valid
normalized range $[0,1]$. The metrics are evaluated on the full images
without boundary shaving and are averaged across all test images. For SSIM,
the spectral bands are treated as image channels and the data range is set
to $1.0$.

\section{Results and Discussion}

\subsection{Quantitative Results}

Experiments are conducted on fixed test sets from SEN2VEN$\mu$S, OLI2MSI, SEN2NAIP, and SpaceNet Challenge 3. Table~\ref{tab:combined_quantitative} compares SFG-SwinSR with bicubic interpolation, CNN-based baselines (EDSR and RCAN), and Swin-based transformer models (Swin2SR and Swin2-MoSE) under identical evaluation settings.

\begin{table*}[!h]
\centering
\caption{Quantitative comparison on SEN2VEN$\mu$S, OLI2MSI, SEN2NAIP, and SpaceNet Challenge 3 datasets. Best results for each metric are shown in bold.}
\label{tab:combined_quantitative}
\small
\setlength{\tabcolsep}{2.5pt}
\begin{tabular}{p{2.8cm} c c c c c}
\toprule
Dataset & Model & Params (M) & PSNR (dB) & SSIM & MAE \tabularnewline
\midrule
SEN2VEN$\mu$S $\times2$ & Bicubic & - & 45.45 & 0.9867 & 0.0036 \tabularnewline
{} & EDSR & 43.10 & 47.62 & 0.9931 & 0.0031 \tabularnewline
{} & RCAN & 15.59 & 48.01 & 0.9926 & 0.0029 \tabularnewline
{} & Swin2SR & 12.09 & 48.43 & 0.9932 & 0.0029 \tabularnewline
{} & Swin2-MoSE~\cite{rossi2025swin2mose} & 11.45 & 48.97 & 0.9937 & 0.0026 \tabularnewline
{} & \textbf{SFG-SwinSR (ours)} & 13.73 & \textbf{49.35} & \textbf{0.9949} & \textbf{0.0025} \tabularnewline
\midrule
SEN2VEN$\mu$S $\times4$ & Bicubic & - & 41.96 & 0.9657 & 0.0059 \tabularnewline
{} & EDSR & 43.10 & 42.58 & 0.9781 & 0.0052 \tabularnewline
{} & RCAN & 15.59 & 43.96 & 0.9771 & 0.0050 \tabularnewline
{} & Swin2SR & 12.09 & 44.12 & 0.9792 & 0.0049 \tabularnewline
{} & Swin2-MoSE~\cite{rossi2025swin2mose} & 11.49 & 45.12 & 0.9806 & 0.0045 \tabularnewline
{} & \textbf{SFG-SwinSR (ours)} & 13.73 & \textbf{45.52} & \textbf{0.9837} & \textbf{0.0042} \tabularnewline
\midrule
OLI2MSI $\times3$ & Bicubic & - & 41.28 & 0.9786 & 0.0064 \tabularnewline
{} & EDSR & 43.10 & 44.31 & 0.9887 & 0.0045 \tabularnewline
{} & RCAN & 15.59 & 44.88 & 0.9896 & 0.0042 \tabularnewline
{} & Swin2SR & 12.09 & 45.32 & 0.9904 & 0.0039 \tabularnewline
{} & Swin2-MoSE~\cite{rossi2025swin2mose} & 11.45 & \textbf{45.94} & 0.9912 & 0.0035 \tabularnewline
{} & \textbf{SFG-SwinSR (ours)} & 13.73 & 45.91 & \textbf{0.9921} & \textbf{0.0032} \tabularnewline
\midrule
SEN2NAIP $\times4$ & Bicubic & - & 32.85 & 0.8914 & 0.0126 \tabularnewline
{} & EDSR & 43.10 & 34.72 & 0.9215 & 0.0103 \tabularnewline
{} & RCAN & 15.59 & 35.14 & 0.9287 & 0.0099 \tabularnewline
{} & Swin2SR & 12.09 & 35.91 & 0.9348 & 0.0094 \tabularnewline
{} & Swin2-MoSE~\cite{rossi2025swin2mose} & 11.45 & 36.42 & 0.9397 & 0.0089 \tabularnewline
{} & \textbf{SFG-SwinSR (ours)} & 13.73 & \textbf{36.95} & \textbf{0.9442} & \textbf{0.0083} \tabularnewline
\midrule
SpaceNet $\times2$ & Bicubic & - & 39.61 & 0.9521 & 0.0048 \tabularnewline
{} & EDSR & 43.10 & 42.37 & 0.9728 & 0.0042 \tabularnewline
{} & RCAN & 15.59 & 43.02 & 0.9756 & 0.0040 \tabularnewline
{} & Swin2SR & 12.09 & 43.56 & 0.9780 & 0.0039 \tabularnewline
{} & Swin2-MoSE~\cite{rossi2025swin2mose} & 11.45 & 44.61 & 0.9816 & 0.0034 \tabularnewline
{} & \textbf{SFG-SwinSR (ours)} & 13.73 & \textbf{45.19} & \textbf{0.9852} & \textbf{0.0031} \tabularnewline
\bottomrule
\end{tabular}
\end{table*}

Overall, SFG-SwinSR provides consistent improvements across most evaluation settings and achieves the best SSIM and MAE across all five evaluation settings. On SEN2VEN$\mu$S, the proposed model improves PSNR over Swin2SR by +0.92 dB and +1.40 dB for $\times2$ and $\times4$ SR, respectively, and also outperforms Swin2-MoSE by +0.38 dB and +0.40 dB. On OLI2MSI, SFG-SwinSR achieves the best SSIM and MAE, while its PSNR remains comparable to Swin2-MoSE, with only a marginal difference of 0.03 dB. On SEN2NAIP, SFG-SwinSR obtains the best performance across all three metrics, improving PSNR by +1.04 dB over Swin2SR and +0.53 dB over Swin2-MoSE. On the auxiliary synthetic SpaceNet Challenge 3 setting, SFG-SwinSR improves over Swin2SR and Swin2-MoSE by +1.63 dB and +0.58 dB, respectively. These results suggest that the proposed spatial-frequency gated FFN provides stable reconstruction gains across real cross-sensor and synthetic SR settings, while remaining a focused refinement of the Swin2SR backbone.

\begin{figure*}[!h]
\centering
\includegraphics[width=\linewidth]{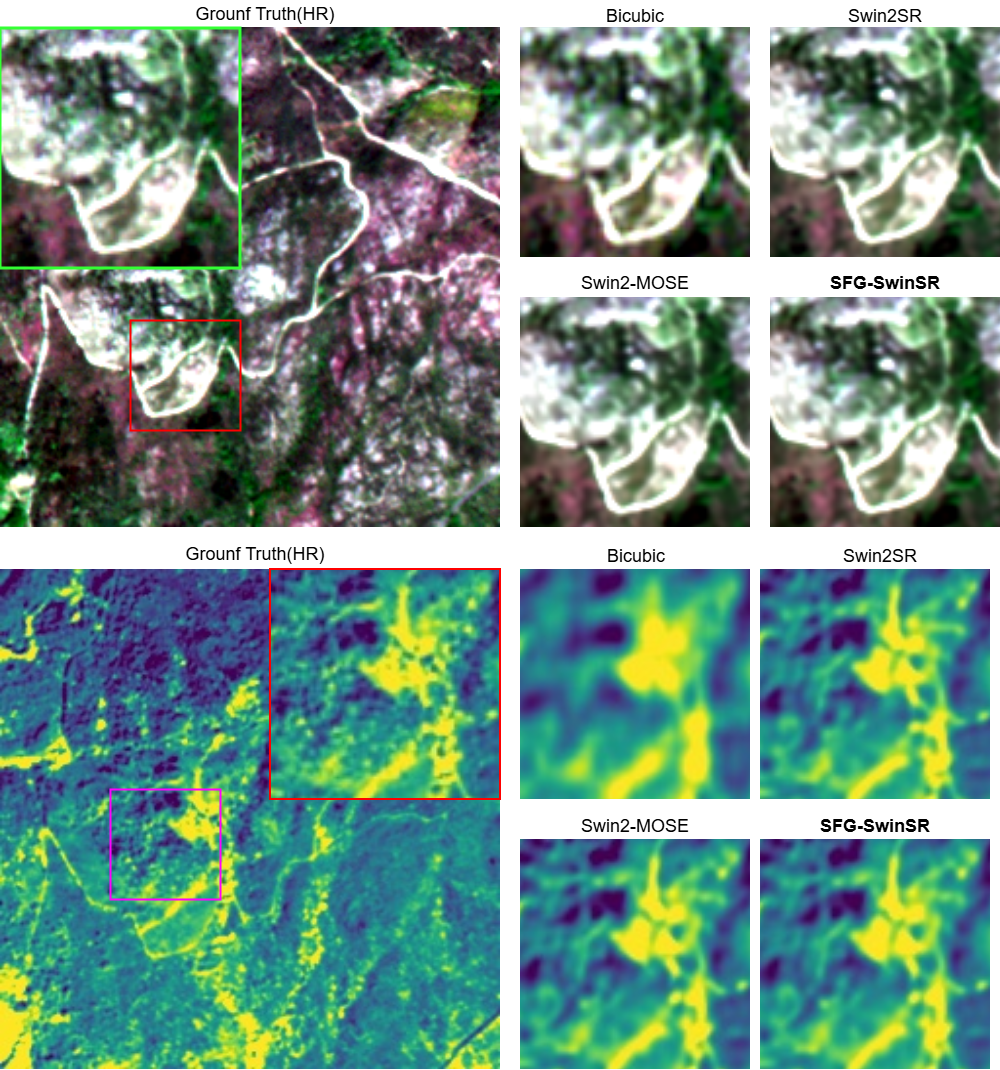}
\caption{Visual comparison on the SEN2VEN$\mu$S dataset. The left panel
shows the HR reference and enlarged region of interest; the right panels
show Bicubic, Swin2SR, Swin2-MoSE, and SFG-SwinSR results. Bounding
boxes indicate the evaluated regions.}
\label{fig:sen2venus}
\end{figure*}

\subsection{Qualitative Analysis}

\begin{figure}[!h]
\centering
\includegraphics[width=\linewidth]{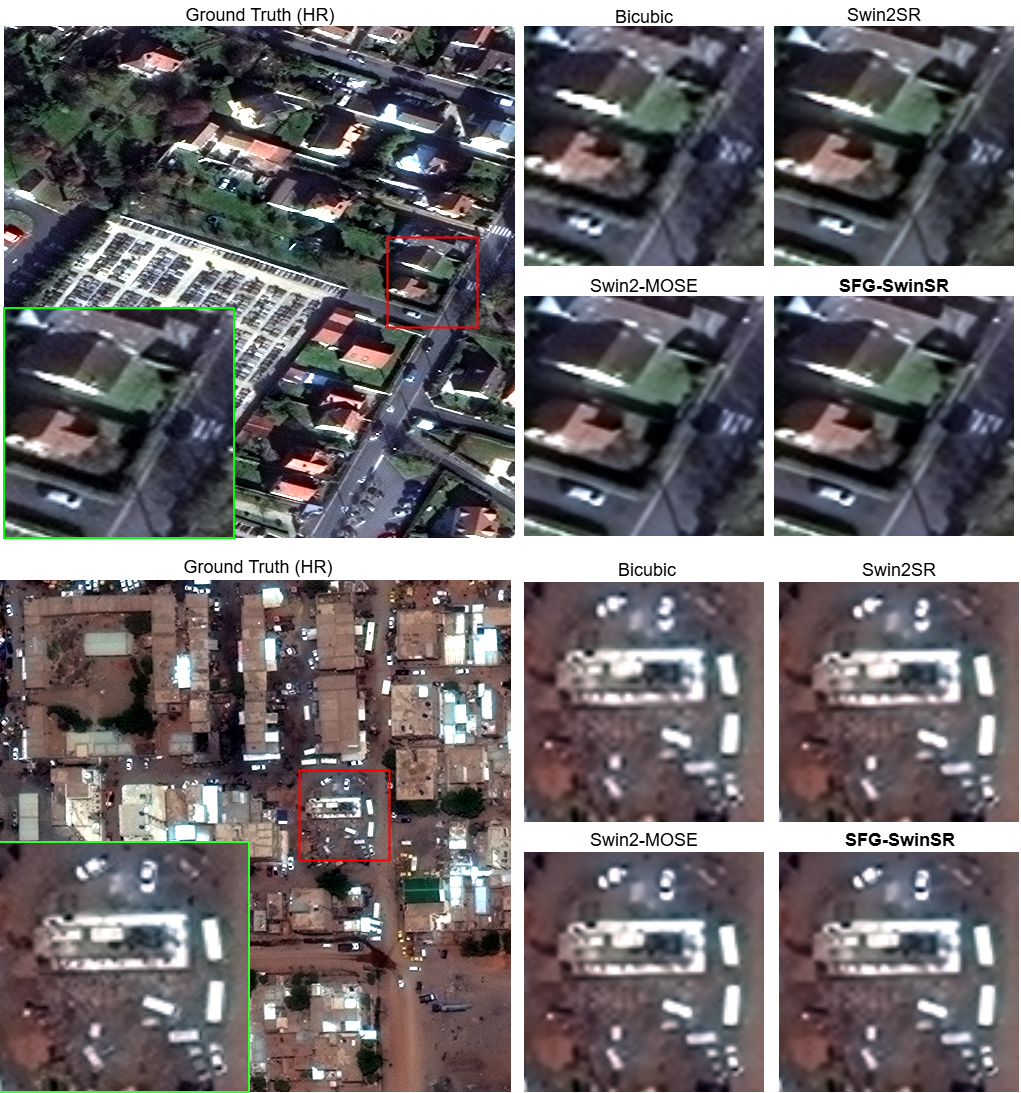}

\caption{Visual comparison on the SpaceNet dataset. The left panels show the ground-truth HR images and enlarged regions of interest, while the right panels show the corresponding Bicubic, Swin2SR, Swin2-MoSE, and SFG-SwinSR results. Green and red bounding boxes indicate the selected regions for qualitative comparison.}

\label{fig:spacenet_visual}
\end{figure}

The qualitative examples in Figures~\ref{fig:sen2venus} and~\ref{fig:spacenet_visual} show that SFG-SwinSR produces sharper structural boundaries and more continuous fine patterns than the baseline models. Improvements are visible around building roofs, road boundaries, and narrow land-cover transitions. At the same time, these visual results should be interpreted as structure-aware reconstruction rather than guaranteed recovery of the true missing high-frequency signal, since single-image SR remains an ill-posed problem.

Overall, the quantitative and qualitative results indicate that the
feature-space low-pass/residual transformation inside the transformer FFN
provides a useful inductive bias for remote sensing SR. The contribution is therefore best understood as a focused and lightweight refinement of Swin2SR, rather than a claim of broad architectural superiority.

\subsection{Ablation Study}

Table~\ref{tab:ablation_study} separates the effects of the backbone and
the training objective on the SpaceNet setting. Under the same complete
training objective, SFG-FFN achieves 45.19~dB compared with 43.86~dB
for the standard FFN, supporting an architectural contribution in this
controlled experiment. Within SFG-FFN, the largest loss-related improvement
is obtained by adding SSIM supervision, which increases PSNR from
42.63~dB to 45.17~dB. The edge and frequency terms provide a smaller
additional improvement from 45.17~dB to 45.19~dB. Therefore, SSIM is the
dominant auxiliary supervision term in this experiment, while the edge and
frequency losses act as incremental refinements.

\begin{table*}[ht]
\centering
\caption{Ablation analysis of the proposed SFG-SwinSR architecture and loss supervision on the SpaceNet dataset.}
\label{tab:ablation_study}
\small
\setlength{\tabcolsep}{6pt}
\begin{tabular}{l c c c c c c c}
\toprule
Backbone Model & L1 & SSIM & Edge & Freq & PSNR (dB) & SSIM & MAE \tabularnewline
\midrule
FFN & \checkmark & & & & 43.56 & 0.9780 & 0.0039 \tabularnewline
FFN & \checkmark & \checkmark & \checkmark & \checkmark & 43.86 & 0.9810 & 0.0039 \tabularnewline
\midrule
SFG-FFN & \checkmark & & & & 42.63 & 0.9785 & 0.0049 \tabularnewline
SFG-FFN & \checkmark & \checkmark & & & 45.17 & 0.9846 & 0.0031 \tabularnewline
SFG-FFN & \checkmark & \checkmark & \checkmark & \checkmark & \textbf{45.19} & \textbf{0.9852} & \textbf{0.0031} \tabularnewline
\bottomrule
\end{tabular}
\end{table*}

Table~\ref{tab:arch_ablation} evaluates the architectural choices
of SFG-FFN on SEN2VEN$\mu$S $\times4$. Under L1-only supervision,
low-pass initialization improves PSNR from 44.78~dB to 44.95~dB relative
to random initialization. This 0.17~dB difference indicates that the
initialization provides a useful prior, although the relatively close results
also show that it is not the sole source of improvement. The $7\times7$
kernel obtains a marginally higher PSNR than the selected $5\times5$
kernel, but uses more parameters. The final row additionally uses the complete
detail-guided objective and is therefore reported as the final system rather
than as an architecture-only comparison.

Because the depthwise low-pass kernel remains trainable, this analysis should
not be interpreted as evidence of an exact physical frequency decomposition.
Instead, the low-pass initialization and residual construction provide a
spatial-frequency inductive bias that encourages the two feature paths to model
different spatial-variation characteristics.

\begin{table}[h]
\centering
\caption{Architectural ablation study of the proposed SFG-FFN on the
SEN2VEN$\mu$S $\times4$ dataset. All architecture-only variants use
$L_1$ supervision, whereas the final SFG-SwinSR row additionally uses
the complete detail-guided objective.}
\label{tab:arch_ablation}
\small
\setlength{\tabcolsep}{3pt}
\begin{tabular}{lcccc}
\toprule
Configuration & Params (M) & PSNR (dB) & SSIM & MAE \tabularnewline
\midrule
Swin2SR Baseline & 12.09 & 44.12 & 0.9792 & 0.0049 \tabularnewline
SFG-FFN w/ random kernel init & 13.73 & 44.78 & 0.9803 & 0.0046 \tabularnewline
SFG-FFN w/ low-pass init & 13.73 & 44.95 & 0.9808 & 0.0045 \tabularnewline
w/o identity HF branch & 13.71 & 44.87 & 0.9805 & 0.0045 \tabularnewline
ReLU instead of GELU & 13.73 & 44.82 & 0.9802 & 0.0046 \tabularnewline
Kernel size $3\times3$ & 13.69 & 44.89 & 0.9804 & 0.0045 \tabularnewline
Kernel size $7\times7$ & 13.79 & 45.01 & 0.9811 & 0.0044 \tabularnewline
w/o DropPath & 13.73 & 44.91 & 0.9807 & 0.0045 \tabularnewline
\textbf{Full SFG-SwinSR} & \textbf{13.73} & \textbf{45.52} & \textbf{0.9837} & \textbf{0.0042} \tabularnewline
\bottomrule
\end{tabular}
\end{table}

\subsection{Parameter and Complexity Analysis}

SFG-SwinSR increases the parameter count from 12.09M for Swin2SR to 13.73M, and the computational cost from 104.06 GFLOPs to 117.23 GFLOPs. This overhead comes from the added depthwise blur branch, spatial refinement branch, and bottleneck gating operation in SFG-FFN. Compared with Swin2-MoSE, SFG-SwinSR has slightly more parameters but
achieves better SSIM and MAE across all five evaluation settings and
competitive PSNR performance, with the best PSNR in four out of five
settings. Thus, the added complexity remains moderate relative to the observed performance gains.

\subsection{Limitations}

This study has several limitations. First, although SFG-SwinSR consistently
improves over the direct Swin2SR baseline, its gains over the stronger
Swin2-MoSE baseline are more modest. In particular, on OLI2MSI,
SFG-SwinSR achieves better SSIM and MAE but remains slightly below
Swin2-MoSE in PSNR, suggesting complementary structural benefits rather
than uniform superiority across all metrics. Second, SpaceNet Challenge 3
is used as an auxiliary synthetic benchmark; its simulated LR--HR pairs
provide a controlled evaluation setting but do not fully capture the
complexity of real cross-sensor observations. Third, the comparison is not
exhaustive with respect to recent generative, diffusion-based, and
degradation-aware SR methods, and the use of PSNR, SSIM, and MAE does
not fully characterize perceptual or spectral fidelity. Finally, because the
depthwise low-pass branch remains trainable, the proposed decomposition
should be interpreted as a spatial-frequency inductive bias rather than an
exact physical frequency separation. Residual-detail refinement may also
amplify noise, haze, clouds, or misregistration artifacts in challenging
cross-sensor scenes. 

Future work will consider broader real cross-sensor evaluation, additional
perceptual and spectral metrics, activation-spectrum analysis, and more robust
degradation-aware frequency gating.
\section{Conclusion}

This paper introduced SFG-SwinSR, a Spatial-Frequency Gated Swin Transformer
for cross-sensor remote sensing image super-resolution. The proposed method
preserves the shifted-window attention mechanism of Swin2SR while replacing
its standard feed-forward network with a lightweight Spatial-Frequency Gated
Feed-Forward Network that estimates a low-frequency feature component, derives
residual detail features, refines them through a spatial branch, and adaptively
reinjects useful details using a bottleneck gate.

Experiments on SEN2VEN$\mu$S, OLI2MSI, SEN2NAIP, and an auxiliary synthetic
SpaceNet Challenge 3 setting show consistent improvements over the direct
Swin2SR baseline and competitive performance against the recent Swin2-MoSE
model, while also outperforming the representative CNN baselines considered
in this study. The method achieves the best SSIM and MAE across all five
evaluation settings and the best PSNR in four out of five settings, with PSNR
gains over Swin2SR of +0.92~dB and +1.40~dB on SEN2VEN$\mu$S $\times2$
and $\times4$, +0.59~dB on OLI2MSI, +1.04~dB on SEN2NAIP, and +1.63~dB
on SpaceNet Challenge 3. These results support the utility of feature-space
spatial-frequency transformation inside transformer FFNs for structure-aware
remote sensing super-resolution.


\bibliographystyle{splncs04}
\bibliography{main}
\end{document}